\documentclass[review]{elsarticle}

\usepackage{lineno,hyperref}
\usepackage{times}
\usepackage{epsfig}
\usepackage{graphicx}
\usepackage{amsmath}
\usepackage{amssymb}
\usepackage{caption}
\usepackage{subcaption}
\usepackage{algorithm}
\usepackage{algorithmic}
\usepackage[toc]{appendix}
\usepackage{xcolor}

\modulolinenumbers[5]
\journal{Journal of Pattern Recognition}

\begin{document}

\begin{frontmatter}

\title{
	Graph Representation Learning for Road Type Classification}

\author[add1]{Zahra Gharaee\corref{mycorrespondingauthor}}
\cortext[mycorrespondingauthor]{Corresponding author}
\ead{zahra.gharaee@liu.se}
\author[add2]{Shreyas Kowshik} 
\ead{shreyaskowshik@iitkgp.ac.in}
\author[add1,add3]{Oliver Stromann}
\ead{oliver.stromann@liu.se}
\author[add1]{Michael Felsberg}
\ead{michael.felsberg@liu.se}
\address[add1]{Computer Vision Laboratory (CVL), Department of Electrical Engineering, University of Link\"oping  \\ Link\"oping, Sweden }
\address[add2]{Department of Mathematics, Indian Institute of Technology Kharagpur, India }
\address[add3]{Autonomous Transport Solutions Research, Scania CV AB, Sweden }

\begin{abstract}
We present a novel learning-based approach to graph representations of road networks employing state-of-the-art graph convolutional neural networks. Our approach is applied to realistic road networks of 17 cities from Open Street Map. While edge features are crucial to generate descriptive graph representations of road networks, graph convolutional networks usually rely on node features only. We show that the highly representative edge features can still be integrated into such networks by applying a line graph transformation. We also propose a method for neighborhood sampling based on a topological neighborhood composed of both local and global neighbors. We compare the performance of learning representations using different types of neighborhood aggregation functions in transductive and inductive tasks and in supervised and unsupervised learning. Furthermore, we propose a novel aggregation approach, Graph Attention Isomorphism Network, GAIN \footnote{Link to repository:https://github.com/zahrag/GAIN}. Our results show that GAIN outperforms state-of-the-art methods on the road type classification problem.  
\end{abstract}


\end{frontmatter}


\section{Introduction}

Cities around the world are growing and increasingly more people are moving from rural to urban areas. Today, 55\% of the world population lives in cities, and by 2050, the number is expected to become 68\% \cite{allison1986urban}. Increased urbanization leads to a stronger need of urban planning and design, which deals with the infrastructure passing into and out of urban areas, such as transportation, communications, and distribution of road networks \cite{allison1986urban}.

In urban planning the physical layout of human settlements is considered as the main subject \cite{taylor1998urban}. Urban design is the process of designing and shaping the physical features of the cities for the provision of municipal services to residents and visitors. In contrast to architecture, which focuses on the design of individual buildings, urban design deals with the larger scale of groups of buildings, infrastructure, streets, public spaces and the whole neighborhoods as well as districts of the entire cities. The main goal is to design urban environments, which are equitable, beautiful, performative, and sustainable \cite{boeing2014leed}.

One key element in designing an urban environment is the design of road networks, which ensures efficient traffic flows as well as connectivity. Road networks design also supports other domains such as autonomous vehicles and entertainment industry for instance gaming.

In this article, we design and implement a graph-based architecture capable of performing the following tasks: \newline  
(1) Learning road networks of realistic cities and towns from open street map to accomplish road type classification.\newline 
(2) Applying line graph transformation to use qualitative road segment features in learning the representations. \newline 
(3) Proposing a neighborhood sampling based on the nodes in local and global topological neighborhoods.\newline
(4) Proposing a novel approach to aggregation, Graph Attention Isomorphism Network, GAIN, and comparing its performance with a variety of state-of-the-art approaches to representation learning of road network graphs.

\section{Related work}

One can find at least three main categories of approaches for modeling road networks in the literature: the earliest are procedural methods for modeling of road networks from a set of rules. Next are example-based approaches applying a preprocessing step to extract statistical information. Most recent are learning-based approaches including methods using deep learning techniques.

The procedural modeling of road networks relies on a set of rules; Open L-System \cite{parish2001procedural} gradually generates a road map from initial seed points to conform user guidance, the automatic procedural road generation \cite{galin2010procedural} uses a weighted anisotropic shortest path algorithm and the procedural modeling in \cite{galin2011authoring} applies a hierarchical generation of road networks from a geometric graph using a non-Euclidean metric combined with a path merging algorithm.

The next category of approaches for modeling road networks covers example-based approaches \cite{emilien2015worldbrush}. In contrast to procedural approaches, example-based methods \cite{nishida2016example} do not utilize a rule set for road generations. They rather analyze the data from road networks or city layouts in a preprocessing step to extract templates as well as statistical information.

The growth of machine learning techniques especially recent advancements in deep learning makes them a powerful tool for road networks learning, prediction and generation. Since road networks are growing extensively, it is required to apply methods capable of dealing with big data. Moreover, access to the satellite images makes it possible to even train deep learning algorithms end-to-end.

\subsection{Learning-based methods}
In recent years deep learning techniques have been used for procedural and data driven content generation. Among them is the method, which learns a low-dimensional generative model from a high-dimensional procedural model by using shape features \cite{yumer2015procedural}. To circumvent the large number of parameters involved in the rule sets and also their non-linear relationship to the resulting content, a sketch-based approach to procedural modeling trains a deep Convolutional Neural Network (CNN) to map sketches into the procedural model parameters \cite{huang2016shape}.

An urban procedural model \cite{nishida2016interactive} also trains CNNs to recognize the procedural rule intended by a sketch and estimating its parameters. They use simple procedural grammars to turn sketches into realistic 3D models. A neurally-guided architecture \cite{ritchie2016neurally} augments procedural models with deep neural networks to control the random selections of different models based on the output.

Using generative models based on deep neural networks have also been extensively studied for graph generations. One example is the method that applies deep learning to generate an initial segmentation of aerial images and feed them to an algorithm, which reasons about missing connections in the extracted road topology \cite{mattyus2017deeproadmapper}.

To automatically construct accurate road network maps from aerial images RoadTracker \cite{bastani2018roadtracer} uses an iterative search process guided by a CNN-based decision function to derive road network graph directly from the output of CNN. Graph neural networks is used to express probabilistic dependencies of a graph nodes and edges in order to learn the distributions over any arbitrary graph \cite{li2018learning}.

The GraphRNN \cite{you2018graphrnn} learns to generate graphs by training on a representative set of graphs and decomposes the graph generation process into a sequence of node and edge formations, conditioned on the graph structure generated so far. The Neural Turtle Graphics (NTG) \cite{chu2019neural} represents the road layout using a graph where nodes of the graph represent control points and edges of the graph represents road segments. NTG is a sequential generative model parameterized by a neural network, which iteratively generates a new node and an edge connecting to an existing node conditioned on the current graph.

There are a number of approaches for graph generation based on Generative Adversarial Networks (GAN). Among them is StreetGAN \cite{hartmann2017streetgan}, in which a preprocessing layer is applied to convert a given representation of a road network into a binary image using pixel intensities to encode the presence or absence of streets. The model next trains a GAN to synthesize a multitude of arbitrary sized street networks. Finally the post-processing layer extracts a graph-based representation of the generated images.

The NetGAN \cite{bojchevski2018netgan} generates graphs from random walks and the model is trained using Wasserstein GAN objective function. Using GANs, GraphGAN \cite{wang2018graphgan} is proposed as a graph representation learning framework, which applies a new graph softmax to satisfy the properties of normalization, graph structure awareness, and computational efficiency.

\subsection{Graph-based methods}
Graphs are more commonly applied to describe information across many diverse fields where complex and unstructured data represents a particular conceptual network like social networks, molecular networks, biological protein-protein networks, telecommunication networks or brain connectomes \cite{velivckovic2017graph}.

The generic structures of such networks, especially when compared to the grid-like structure of the images, audio, and text, makes it difficult to analyze their representations. Since graphs are non-Euclidean and the number of vertices and edges can vary arbitrarily, they become a powerful tool for data representation in such irregular domains and therefore, it has been a surge in research on graph representation learning recently \cite{hamilton2017representation}.

To incorporate high-dimensional non-Euclidean information of the graph structure, earlier approaches relied on hand-engineered features such as degrees or clustering coefficients \cite{bhagat2011node}, kernel functions \cite{vishwanathan2010graph} and engineered features to measure local neighborhood structures \cite{liben2007link}. However, using hand-engineered features could be an inflexible and expensive task.

Recently learning representations is becoming more popular, which is based on learning a mapping that embeds nodes, or the entire (sub)graphs, as points in a low-dimensional vector space to summarize every node's position and the structure of its local neighborhood \cite{hamilton2017representation}. The learning representations facilitates downstream machine learning tasks like link prediction \cite{berg2017graph} or node classification \cite{hamilton2017inductive}.

Graph structure and methods have also been used to enhance the ability to learn a variety of tasks like image clustering and recognition \cite{Xiao2009}, object identification \cite{Xiao2011}, action recognition \cite{CHEN2020107321}, and few-shot learning (FSL) \cite{DBLP:conf/iclr/SatorrasE18}. The approach to image clustering and recognition \cite{Xiao2009} investigates graph characteristics from the heat kernel trace by exploring three different methods to characterizing it as a function of time. To identify objects, spectral graph analysis of a hierarchical description of an image is applied to construct a feature vector of fixed dimension \cite{Xiao2011}.

Among the graph-based methods developing FSL are the EGNN \cite{DBLP:conf/cvpr/KimKKY19}, which learns to predict edge-labels rather than node-labels by iteratively updating the edge-labels using both intra-cluster similarity and the inter-cluster dissimilarity, and the DPGN \cite{DBLP:conf/cvpr/YangLZZZL20}, which incorporates distribution propagation in graph neural network to facilitate FSL tasks. In \cite{ZHANG2021107946}, however, a concept graph is used for weakly supervised FSL by applying a meta concept inference network, which quickly adapts to a novel task using the joint inference of the abstract concepts and a few annotated samples.

\subsection{Graph convolutional networks}

Graph Convolutional Networks (GCN) are a recent generation of approaches, which represent every node of a graph as a function of its neighborhood \cite{kipf2016semi, kipf2016variational, berg2017graph}. Learning node representations has three important advantages. First is its computational efficiency as a result of sharing network parameters, second is the integration of nodes attributes to generate information about their positions and roles in the graph, and third is the generalization of learned knowledge to the unseen nodes and graphs.

Aggregating the information of a local node neighborhood instead of the entire graph will therefore help to address the limitations of approaches relying on shallow embedding \cite{perozzi2014deepwalk, grover2016node2vec} and the ones using deep auto-encoders \cite{cao2016deep, wang2016structural}.

Similar to GCN, GraphSAGE \cite{hamilton2017inductive} applies convolutional mechanism to train a set of aggregator functions in order to learn aggregated information of a local neighborhood. GraphSAGE can leverage node features in order to learn an embedding function, which generalizes to unseen nodes or graphs. However, GraphSAGE and GCN differ in the aggregation as well as vector combination methods they apply.

An attention based graph convolutional approach, Graph Attention Network (GAT)  \cite{velivckovic2017graph} operates on graph-structured data using masked self-attention layers to improve the performance of prior graph convolution based methods. The gated attention network (GaAN) \cite{zhang2018gaan}, on the other hand, applies a convolutional sub-network to control the importance of attention head based on a number of gates.

The more recent, Graph Isomorphism Network (GIN) \cite{xu2018powerful} models injective multiset functions for neighborhood aggregation by developing a theory of “deep multi-sets”, which parametrizes universal multi-set functions using neural networks such as multi-layer perceptrons, MLP. GIN applies $\textsc{{SUM}}$ aggregators to implement injective and universal functions over the multi-sets.

Using recent graph representation learning approaches such as GraphSAGE \cite{hamilton2017inductive}, GAT \cite{velivckovic2017graph}, GaAN \cite{zhang2018gaan}, and GIN \cite{xu2018powerful}, a node learns representation by aggregating the information of nodes sampled from its local neighborhood through a certain number of hops, where one hop is moving one layer forward from a node.

Learning graph structure using local neighborhood aggregation, however, is not capable of integrating edge features in graph representation learning as algorithms rely on node features only. However, in road network graphs, edge features are more descriptive and could play a significant role in learning representations. To address this issue, representation fusion of nodes, edges, and between-edges, the Relational Fusion Network (RFN) cite{jepsen2020relational} for road network graphs is proposed. RFN uses both primal and dual graphs where dual graph nodes and edges represent primal graph edges and between-edges, respectively. Applying relational fusion, RFN addresses speed limit classification and speed limit estimation problems in an inductive supervised setting using binary classification.

Similar to RFN \cite{jepsen2020relational}, we apply dual graph in our experiments, however, we use exclusively dual graph generated by line graph transformation of the original graph in order to make use of informative road segment attributes in learning representations and, therefore, we do not make use of original graph in our analysis. To the best of our knowledge, the basic approach to generate line graph in this article is similar to the one used in dual graph presented in \cite{jepsen2020relational}. 

Moreover, in contrast to RFN, which only addresses speed limit classification and estimation in an inductive supervised setting using binary classes, our experiments explore the four major tasks of multi-class road type classification in unsupervised and supervised transductive as well as inductive settings. A more detailed description of experimental setup used by RFN in comparison to ours is presented in section \ref{sec:rfn}.

The unsupervised classification of road networks is an important problem since completing missing labels is logistically demanding and expensive task. In OSMnx \cite{boeing2017osmnx}, labels are frequently left out or road types are miss-labeled. Especially the inductive setting allows training on high-quality densely labeled road network and transferring this knowledge onto more sparsely labeled road networks.

Furthermore, this article proposes a novel approach, GAIN to aggregating knowledge for learning representations used to address road type classification problem. However, GAIN could be applied to learning representations of any types of graphs and not just road type graphs and, therefore, any classification task could be addressed using the same approach.

To enhance learning representations, we also propose sampling from a topological neighborhood composed of both local (GraphSAGE \cite{hamilton2017inductive}) and global neighbor nodes by applying a new search mechanism presented in Section \ref{sec:topo-neigh}. This facilitates using relevant information of the neighbor nodes in further distances to a node.

\section{Method}

In a standard graph learning problem, a graph is represented by a set of nodes or the vertices connected by a set of links or the edges. This formulation is  usually denoted by $G=(V, E)$, where $G$ is the graph, $V$ and $E$ are vertices and edges respectively. In this article, the learning paradigm starts by setting a number of hops, where one hop is counted by moving forward one layer from a node. The number of hops is set to 2 and the graph node attributes are represented by the nodes in the deepest hop.

Sampling among nodes of the training set, the algorithm selects two sets of nodes, one set per hop. The second set contains nodes in the neighborhood of the first set. For every node of the second set, it aggregates information of direct neighbors to its own and, therefore, it generates a vector representing itself. Similar to the second set, each node of the first set generates its representation vector by aggregating information from direct neighbors to its own.

\subsection{Line graph transformation}\label{linegraph}
\begin{figure}[h]
	\centering
	\includegraphics[width=1\linewidth]{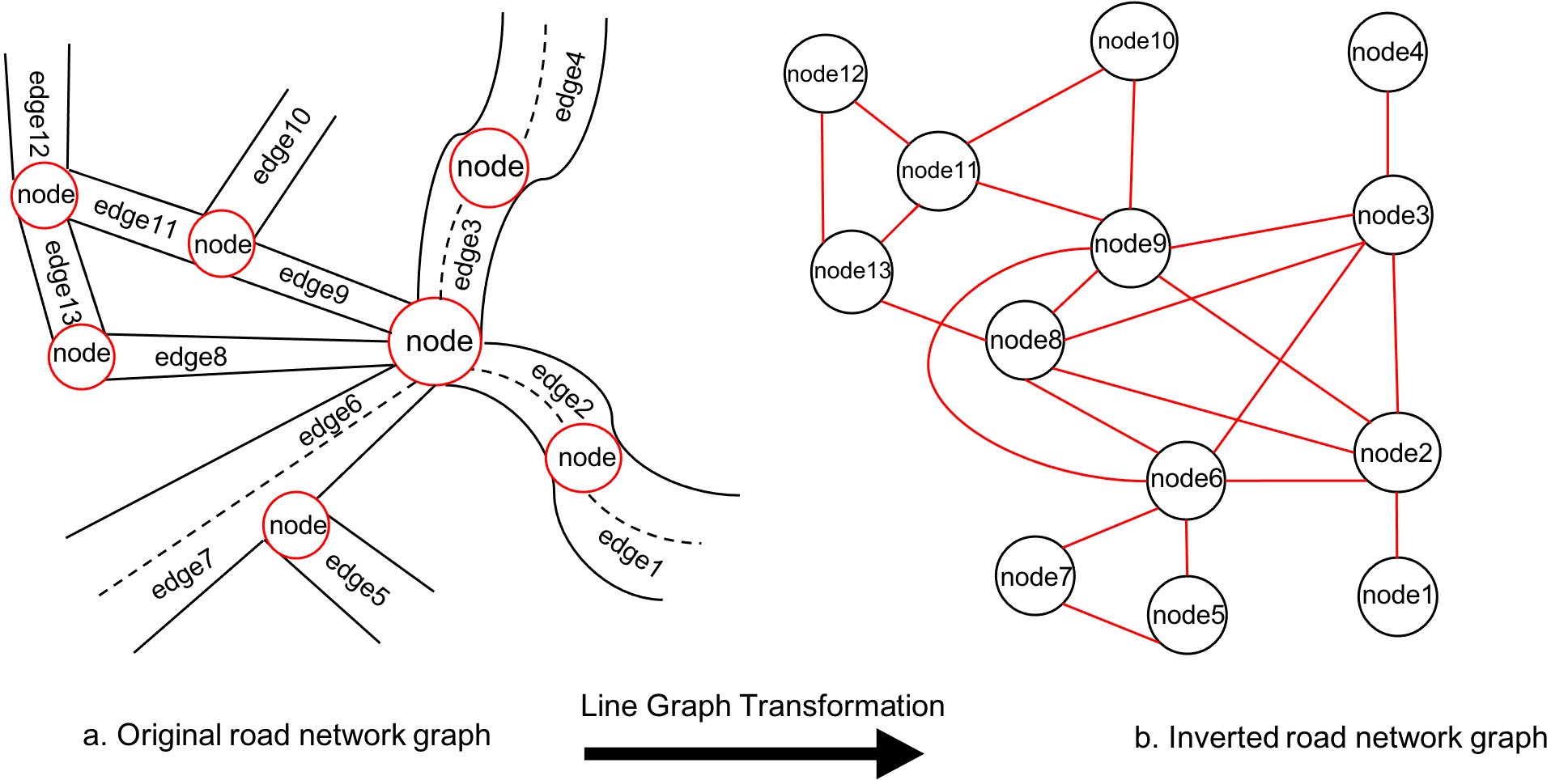}
	\caption{A sketch of applying the line graph transformation on a road network graph data (a), which results in a new graphical network (b).}
	\label{fig:line-graph-T}
\end{figure}

Having a first view to the road network graphs, it sounds reasonable to consider road segments as graph edges and crossroads, junctions, and intersections as graph vertices as shown in figure \ref{fig:line-graph-T} (a). However, this approach suffers from a limited feature representation of vertices since there are not sufficient features describing crossroads and intersections that are essential for road network representation.

Furthermore, among many approaches addressing the graph representation learning problem \cite{hamilton2017inductive, velivckovic2017graph, xu2018powerful, zhang2018gaan}, the features used in the learning process describe merely the vertices and not the edges.

To address feature representation problem, we instead replaced graph vertices as roads segments and, therefore, the connectivity between road segments will generate graph edges as shown by figure \ref{fig:line-graph-T} (b). Using this strategy, the informative road features can be utilized in the learning representation process. A more detailed description about implementation of line graph transformation is available in section \ref{sec:input}.

\subsection{Supervision modes}
\paragraph{Supervised learning} For supervised learning, ground-truth road type labels of graph vertices are used for calculating a cross-entropy loss function for our multi-class road type classification problem. The representation vectors received from the aggregation function are normalized by $l_2$ normalization and input to a one-layer fully connected neural network to predict class labels, which are then used to calculate the supervised loss value.

\paragraph{Unsupervised learning}
A fully unsupervised setting uses a graph-based loss function shown in (\ref{eq:1}) to the positive case representation vectors sampled from a set of topological neighbors and negative case sampling distribution, $P_n$:

\begin{equation}\label{eq:1}
J_G(\textup{z}_v) = - \text{log} (\sigma(\textup{z}_v^T\textup{z}_u)) - \mathop{\mathbb{E}}_{u_n\sim P_n(u)}\text{log} (\sigma(-\textup{z}_v^T\textup{z}_{u_n})),
\end{equation}

where $z_v$ and $z_u$ are the output representation vectors of sampled node $v$ and topological neighbor node $u\in N_t(v)$. The loss function uses a sigmoid $\sigma$ and negative case sampling distribution $P_n$. Thus, $\textup{z}_{u_n}$ is a negatively sampled neighbor of node $v$.   

\subsection{Building topological neighborhoods}\label{sec:topo-neigh}
We propose a set of topological neighborhoods, $N_t(v)$ for every sampled node $v$ of the Graph, which encompasses both local, $N_l(v)$, and global, $N_g(v)$, neighborhoods. To this end, a search mechanism is implemented, which selects local and a set of global neighbors for a node $v$ based on an unbiased random walk as presented by Algorithm \ref{alg:alg1}.

Applying both local and global neighborhood provides a node with two views. Having the first view, a node captures the closer vicinity and it extracts the information of its neighbors in a fixed length local area \cite{hamilton2017inductive}. We propose adding a second view based on the node global neighborhood to facilitate the extraction of information from related nodes in further distances. This way, we extend the node representation to improve the learning procedure.

To generate the second view as mentioned above, we developed a global random walk with fixed length, $L_g$ two times the size of local random walk, $L_l$ as $L_g=2 \times L_l$. All topological neighbors visible to a node through both views are then mixed and shuffled to be used for training the system.

\begin{algorithm}
	\caption{Topological Neighborhood: $N_t(v)$ }
	\label{alg:system}
	\textbf{Input}: $G=(V, E)$ \& $N(v\in V)=\{u|(v, u)\in E\}$.\\
	\textbf{Output}: Topological neighborhood $N_t(v)$ of local $N_l(v)$ \& global $N_g(v)$ neighbors.\\
	\vspace{-2em}
	\begin{algorithmic}[1] 
		\STATE Initialize number of walks $N_w$, local walk length $L_l$ \& global walk length $L_g$.
		\FOR {$v \in V$}
		\STATE $N_l(v)=N_g(v)=N_t(v)=\{\}$.
		\FOR {$n=1:N_w$ }
		\FOR {$l=1:L_l$ }
		\STATE Sample $u$ from $N(v)$. 
		\IF {$u \neq v$ }
		\STATE  $N_l(v) \leftarrow u$.
		\ENDIF
		\ENDFOR \\
		\STATE $u_1=v$.\\
		\STATE Sample $u_2$ from $N(v)$.
		\FOR {$l=1:L_g$ }
		\STATE $u_0 = u_1$.
		\STATE $u_1 = u_2$.
		\STATE Sample $u_2$ from $N(u_1)$.
		\ENDFOR
		\IF {$u_1 \neq v$ }
		\STATE  $N_g(v) \leftarrow u_1$. 
		\ENDIF	 
		\ENDFOR \\
		$N_t(v) = N_l(v) \cup N_g(v)$	
		\ENDFOR
	\end{algorithmic}\label{alg:alg1}
\end{algorithm}

\subsection{Previous approaches proposed for aggregation}
As mentioned earlier, a sampled node $v$ aggregates the information of its direct neighbors $u \in N(v)$ in order to generate an output representation vector $\textup{{h}}^k_v$ for each hop layer $k$. There are a number of different approaches used to build the aggregation function, which are mentioned in the following.  

\paragraph{GCN}
A graph convolutional network, GCN \cite{kipf2016semi} aggregates the information as:

\begin{equation}\label{eq:2}
\textup{{h}}_v^k \leftarrow \sigma\left({\textup{{W}}}\cdot \textsc{{MEAN}}\left( \textup{{h}}_v^{k-1}\mathbin\Vert \textup{{h}}_{u \in N(v)}^{k-1}\right)\right),
\end{equation} 

where $\textup{{W}}$ is the set of weights, associated to the sampled and the neighbor nodes represented by $v$ and $u$. $k$ iterates over hop layers and $\sigma$ is the sigmoid function. Concatenation shown by $\mathbin\Vert$ is applied to the sampled and neighbor nodes representation vectors before applying the \textsc{{MEAN}} operation.

\paragraph{GraphSAGE}
GraphSAGE architecture \cite{hamilton2017inductive} applies a different formulation to aggregation:

\begin{equation}\label{eq:3}
\textup{{h}}_v^k \leftarrow \sigma\left({\textup{{W}}}\cdot \left(\textup{{h}}_v^{k-1}\mathbin\Vert \textsc{AGG} \left(\textup{{h}}_{u \in \mathcal{N}(v)}^{k-1}\right)\right)\right),
\end{equation}

where $\textup{{W}}$ is the set of weights, associated to the sampled and the neighbor nodes represented by $v$ and $u$, and $\sigma$ is the sigmoid function. For each hop $k$, the representation vectors of neighbors $\textup{{h}}_{u \in N(v)}^{k-1}$ of the sampled node $v$ are aggregated using an aggregation function $\textsc{AGG}$ and then the aggregated vector is concatenated to the representation vector of the sampled node $\textup{{h}}_v^{k-1}$.

To make a thorough analysis, we made our investigations based on a number aggregation functions \cite{hamilton2017inductive}: GraphSAGE-MEAN, which applies mean operation to aggregate neighborhood information, GraphSAGE-MEANPOOL, which aggregates information by calculating mean pooling over $\textsc{{MLP}}$ functions, GraphSAGE-MAXPOOL, which aggregates information using max pooling of neighborhood information over $\textsc{{MLP}}$ functions and, finally, GraphSAGE-LSTM, which applies a standard \textsc{{LSTM}} to aggregate the neighborgood information. 

\paragraph{GAT} The graph attention network, GAT \cite{velivckovic2017graph}, calculates average of the weighted representation vectors of the first order neighbor nodes $u \in N(v)$ (including $v$) over multiple heads by applying attention weights, $\alpha^m_{vu}$, to the corresponding neighbors:
 
\begin{equation}\label{eq:4}
\textup{{h}}_v^k = \sigma\left(\frac{1}{\textup{{M}}}\sum_{m=1}^{M}\sum_{u\in N(v)}\alpha^{k-1}_{m_{vu}}\textup{{W}}^{\prime}_{m} \textup{{h}}_u^{k-1}\right),
\end{equation}

where $\textsc{{M}}$ is the total number of attention heads used for regularization and $\textup{{W}}^{\prime}$ is a set of corresponding input linear transformation weights matrix used to project nodes input features into a higher level feature space.

\paragraph{GIN}The graph isomorphism network, GIN \cite{xu2018powerful} models injective multi-set functions for aggregating information by parameterizing multi-layer perceptrons, $\textsc{{MLP}}$. GIN employs the aggregation formulation as:  

\begin{equation}\label{eq:5}
\textup{{h}}_{v}^{k} \leftarrow \textsc{{MLP}}^{k} \left((1+\epsilon^k) \cdot \textup{{h}}_v^{k-1}+\sum_{u\in N(v)} \textup{{h}}^{k-1}_u\right),
\end{equation}

where $\textsc{{MLP}}$ is the multi-layer perceptron and $\epsilon$ could be either learned by gradient descent as one variable of the network or be fixed to zero.

\subsubsection{Novel formulation of the aggregation function}
As an alternative to the explained aggregation methods, we developed a novel formulation of aggregation, the Graph Attention Isomorphism Network (GAIN) \footnote{Isomorphism term in GAIN is used as a reference to GIN representing the bases of our approach, however our proposed approach to aggregation does not fully hold injective features due to the application of non-linearity in $\sigma$ and attention weights.}. Using this approach, a node aggregates information of its neighbors based on an importance value given to each neighbor node and applies the $\textsc{{SUM}}$ for aggregation:

\begin{equation}\label{eq:6}
\textup{{h}}_{v}^{k} = \textsc{{MLP}}^{k} \left((1+\epsilon^k)\cdot \textup{{h}}_v^{\prime k-1} + \sigma \sum_{u \in N(v)} a_{{v, u}}^{k-1} \cdot \textup{{h}}_{u}^{\prime k-1}\right),
\end{equation}

where $\textup{{h}}_{u}^{\prime}=\textup{{W}}^{\prime}_m \cdot \textup{{h}}_{u}$ shows the linear transformation of node $u$ into a higher level feature space using weight matrix $\textup{{W}}^{\prime}_m$. The attention weight $a_{{v, u}}$ is given to the neighbor node $u\in N(v)$.

In our implementations, we applied one $\textsc{{MLP}}$ function with one hidden layer since an $\textsc{{MLP}}$ can represent the composition of functions \cite{weisfeiler1968reduction}. We implemented non-linearity $\sigma$ using both $\textsc{{ELU}}$ and Identity function, and we observed that the performance was slightly superior applying Identity function. 

Inspired by GAT \cite{velivckovic2017graph}, a feed-forward neural network with weights $\textup{{W}}_a$ is applied to the concatenation of sampled and neighbor nodes. This concatenated vector is then given to a non-linear leaky $\textsc{{RELU}}$ function:

\begin{equation}\label{eq:7}
\hat{a}_{w_{v, u}} = \textsc{{relu}}\left(\textup{{W}}_{a} \cdot \left(\textup{{h}}_v^{\prime k-1}\parallel_{u\in N(v)} \\ \textup{{h}}_u^{\prime k-1}\right)\right). 
\end{equation}

Finally, a soft-max function is used to create attention weights:

\begin{equation}\label{eq:8}
a_{{v, u}} = \frac{\exp(\hat{a}_{w_{v, u}})}{\sum_{u\in N(v)} \exp(\hat{a}_{w_{v, u}})}. 
\end{equation}

Compared to GAT \cite{velivckovic2017graph}, GAIN applies attention weights only to the neighboring nodes of $v$ (excluding $v$) and it uses $\textsc{{SUM}}$ rather than $\textsc{{MEAN}}$ over attention heads. On the other hand, GAIN applies attention weights to aggregate neighborhood representations rather than just aggregating neighborhood representations without weighting them as proposed by GIN \cite{xu2018powerful}.

GAIN formulation presented in (\ref{eq:6}) considers one attention head, however, the mathematical formulation of GAIN using multi-head attention is presented in Appendix \ref{sec:app1}. Our experimental investigation shows no improvement in performance applying multi-head attention (\ref{eq:4}).

\section{Experiments}

To evaluate the ability to generalize to unseen graphs, we designed experiments using two different data sets, inductive vs. transductive setting, which denotes inductive reasoning as inferring knowledge from specific to general and transductive reasoning as inferring knowledge from specific to specific \cite{hamiltonbook}. 

Hence, for the inductive experiments we report the results of learning representations on test sets containing graphs of unseen cities and for the transductive experiments we report the result of learning representation on unseen test nodes of the same graph. Another split of our experiments is designed to compare the performance of unsupervised vs. supervised training.

\begin{figure}[h]
	\centering
	\includegraphics[width=1\linewidth]{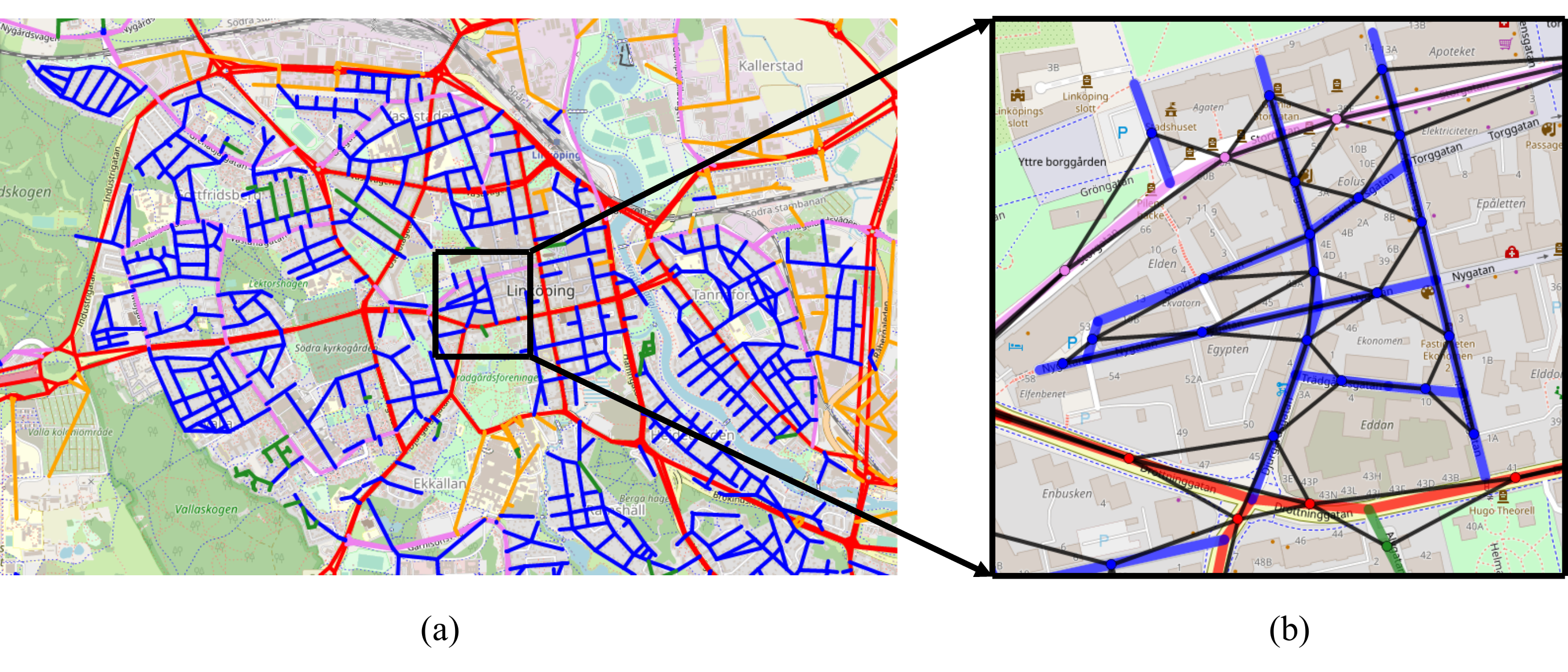}
	\caption{Input data generation as road network graphs. (a) Link\"oping road networks represented as a graph. Colors represent different ground truth labels of road types. (b) A closeup of Link\"oping road networks represented as a graph with a line graph representation overlaid in black. Colors represent different ground truth labels of road types.}
	\label{fig:fig2}
\end{figure}

\subsection{Input Dataset} \label{sec:input}
To address the main problems of this project, road network datasets are represented as graphs composed of vertices and edges. To test transductive and inductive capabilities of the assessed methods, we generate two datasets of road networks. Using Open Street Map (OSMnx \cite{boeing2017osmnx}), we extract the crowd-sourced geographic information of road networks in Swedish Cities from OSMnx.

Both datasets are preprocessed in the following way. The OSMnx data of driving roads is extracted from a 14 km $\times$ 14 km tile centered at the city centroid. The resulting graph is simplified such that intersections are consolidated within a 10 m distance and interstitial nodes are reduced. Directions of edges are removed and parallel edges are consolidated, a limitation necessary to apply the graph representation learning methods.

We convert graph $G$ into a line graph $L(G)$, as described in Section \ref{linegraph}. Each edge of $G$ becomes a node in $L(G)$ and two edges that share a common node in $G$ become an edge in $L(G)$. Figure \ref{fig:fig2} illustrates edges of the original graph $G$ colored by their different ground truth road type labels. Overlaid is the line graph representation with black edges and colored nodes corresponding to road type label.

The transductive line graph consists of the road network of Link\"oping (population of 106,502) with 500 nodes held out for validation and 1000 nodes for testing. The inductive line graph consists of disjoint road networks of 17 Swedish Cities with populations between 50.000 and 150.000 inhabitants. We excluded the three major cities Stockholm, G\"oteborg and Malm\"o as well as any suburbs of them, as they have different magnitudes of populations and exhibit different road network characteristics. Two cities each were held-out for validation and testing respectively and, therefore, 13 graphs are allocated to training the network. Table\ref{tab:tab1} provides a summary of the transductive and inductive datasets.

\subsubsection{Raw features} \label{sec:feat}
To address the road-type classification problem, we extracted descriptive attributes of road-segments represented by the edges of the original graph, $G$ as well as the nodes of the transformed graph, $L(G)$. These attributes are used to create raw feature vectors and in the graph representation learning task. In general the feature vector is composed of 4 main component described as:
\begin{itemize}
	\item The length of road segments with 1 dimension.
	\vspace{-0.6em}
	\item The midpoint coordinates of adjacent start and end nodes in longitude and latitude with 2 dimensions.
	\vspace{-0.6em}
	\item The geometry sampled to a fixed-length vector of 20 equally distanced points along the length of road segments, which is subtracted by the midpoint coordinate (i.e. 20 longitudinal and latitudinal distances from the midpoint) and is composed of 40 dimensions.
	\vspace{-0.6em}
	\item The one-hot-encoding of the speed limits with 13 and 15 standard values for transductive and inductive tasks, respectively.
\end{itemize}

As a consequence, the raw feature vectors of roads segments is composed of 56 and 58 values for the transductive and inductive tasks, respectively.

\subsubsection{Ground-truth labels}\label{sec:inputgt}
To accomplish supervised experiments, it is required to have graph vertices annotated using ground-truth labels. In OSMnx, roads are tagged with road type labels applicable for the classification tasks. However, due to extreme class imbalances shown by figure \ref{fig:fig_}, some classes rarely occurring in our data set. Therefore, we chose to merge and relabel classes according to the following scheme:

\begin{figure}[h]
	\centering
	\includegraphics[width=1\linewidth]{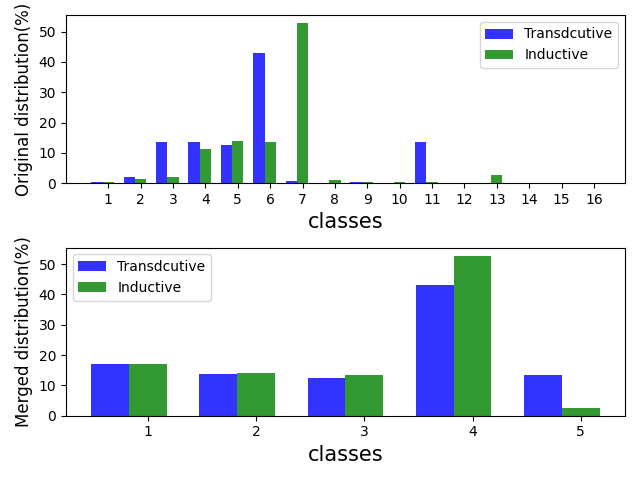}
	\caption{
			Percentage of road network class distribution for transductive and inductive data sets generated by OSMnx \cite{boeing2017osmnx}. As shown by the upper image the class distribution is extremely unbalanced and, therefore, we merged classes to get a more balanced distribution of class labels shown by the bottom image.}
	\label{fig:fig_}
\end{figure}

\begin{table*}
	\begin{center}
		\begin{tabular}{|l| c| c| }
			\hline
			\multicolumn{3}{|c|}{\textbf{Road Network Datasets}} \\
			\hline 
			\textbf{Task}                  & Transductive  & Inductive\\ 
			\textbf{\# Graphs}             & 1             & 17\\
			\textbf{\# Nodes}              & 6903          & 66580\\
			\textbf{\# Edges}              & 13199         & 128632\\
			\textbf{\# Raw Features}       & 56            & 58 \\
			\textbf{\# Road-Type Classes}  & 5             & 5 \\
			\textbf{\# Training Nodes}     & 5403          & 53494\\
			\textbf{\# Validation Nodes}   & 500           & 6575\\
			\textbf{\# Test Nodes}         & 1000          & 6511\\
			\textbf{Avg. Node Degree}      & 3.8241        & 3.8640 \\
			\hline
		\end{tabular}
		\caption{Summary of data sets created for our experiments.}
		\label{tab:tab1}
	\end{center}
\end{table*}

\begin{itemize}
	\item Class (1): Highway, yes, primary, secondary, motorway-link, trunk-link, primary-link, secondary-link.
	\vspace{-0.6em}
	\item Class (2): Tertiary, tertiary-link.
	\vspace{-0.6em}
	\item Class (3): Road, planned, unclassified (minor roads of lower classification than tertiary).
	\vspace{-0.6em}
	\item Class (4): Residential.
	\vspace{-0.6em}
	\item Class (5): Living-street.
\end{itemize}

\begin{table}[h]
	\begin{subtable}[h]{0.45\textwidth}
		\centering
		\small
		\begin{tabular}{|p{0.57\textwidth} | l | l|}	                            
			\hline 
			\textbf{Approach}      & \textbf{Unsup}. & \textbf{Sup}.  \\
			\hline
			\textbf{Random Baseline}                & 0.20            & 0.20\\
			\textbf{Raw Features}          & 0.59            & 0.59\\
			\hline
			\textbf{GCN}                   & 0.60            & 0.58\\
			\textbf{GSAGE-MEAN}            & 0.67            & 0.62\\
			\textbf{GSAGE-MEANPOOL}        & 0.69            & \textbf{0.81}\\
			\textbf{GSAGE-MAXPOOL}         & 0.68            & 0.80\\
			\textbf{GSAGE-LSTM}            & 0.69            & \textbf{0.81}\\
			\textbf{GAT}                   & 0.69            & 0.75\\
			\textbf{GIN}                   & 0.69            & 0.78\\
			\textbf{GAIN (ours)}           & \textbf{0.71}   & \textbf{0.81}\\
			\hline
			\textbf{\%gain over Baseline }     & 20\%        & 37\%  \\
			\hline                
		\end{tabular}
		\caption{Transductive-Task}
		\label{tab:Transductive-Task}
	\end{subtable}
	\hspace{1.5em}
	\begin{subtable}[h]{0.45\textwidth}
		\small
		\begin{tabular}{|p{0.57\textwidth} | l | l|}
        \hline 
        \textbf{Approach}      & \textbf{Unsup}. & \textbf{Sup}.  \\
        \hline
			\textbf{Random Baseline}                    & 0.20             & 0.20\\
			\textbf{Raw Features}                   & 0.49             & 0.49\\
			\hline
			\textbf{GCN}                   & \textbf{0.61}    & 0.43\\
			\textbf{GSAGE-MEAN}            & 0.60             & \textbf{0.60}\\
			\textbf{GSAGE-MEANPOOL}        & \textbf{0.61}    & 0.45\\
			\textbf{GSAGE-MAXPOOL}         & 0.55             & 0.44\\
			\textbf{GSAGE-LSTM}            & 0.59             & 0.45\\
			\textbf{GAT}                   & 0.51             & 0.43\\
			\textbf{GIN}                   & 0.47             & 0.46\\
			\textbf{GAIN (ours)}           & 0.56             & \textbf{0.59}\\
			\hline
			\textbf{\%gain over Baseline}      & 24\%             & 22\% \\
			 \hline  
		\end{tabular}
		\caption{Inductive-Task}
		\label{tab:Inductive-Task}
	\end{subtable}
	\caption{
			Performance of unsupervised and supervised graph representation learning networks accomplished the transductive (a) and inductive (b) tasks represented by micro-averaged F1 Scores on the road type classification problem. The results are compared with both random baseline performance and when only the raw features are given to the classifier. Gain over baseline computes the percentage of improvement to when using raw features only. The results show a challenging classification problem mainly due to the imbalances apparent in the class distribution presented in section 4.2.1.}
	\label{tab:tab2}
\end{table}

\subsection{Results}
The input dataset of road networks graphs described in section \ref{sec:input} are then used to train the graph representation learning algorithm using 8 different aggregation functions. The experiments are designed to investigate performances of different aggregation functions first when learning supervised vs. unsupervised and then when performing an inductive vs. a transductive task.

As shown in table \ref{tab:tab2}, the experiments are conducted by 8 different graph representation learning approaches using ADAM optimiser through an exhaustive grid search over a set of different learning rates and output dimensions. The best performing model for each approach on the validation set in terms of micro-averaged F1-Score is selected and tested on the test set.

\subsubsection{Hyperparameter settings}
\label{sec:paramset}

The settings of all hyperparameters required to run the experiments are presented in table \ref{tab:tab3}. The learning rate and output dimensions are used to conduct exclusive grid search to find the best performing model based on the validation set.

\paragraph{Unsupervised} 
For all unsupervised experiments, an exhaustive grid search is conducted over the learning rates of $\{2e^{-8}, 2e^{-7}, 2e^{-6}, 2e^{-5}\}$ and the output dimensions of $\{64, 128, 256\}$ for the representation vectors at every depth $k$ of the recursion. There are 9 and 3 neighbors sampled for aggregation in the first and second hop layers, respectively. There are 12 negative neighbors sampled and selected for the unsupervised graph-based loss function shown by (\ref{eq:1}) and a dropout rate of 0.1 is used. We have also made an exhaustive grid search to study the role of $\epsilon$ in GAIN (\ref{eq:6}) by first fixing it to zero and then as a variable of the network learned by gradient descent initialized by 0.001 and 0.5. Further investigations about the role of $\epsilon$ could be conducted in future studies.
 
\begin{figure}
	\begin{subfigure}{.5\textwidth}
		\centering
		\includegraphics[width=5.4cm,height=4.1cm]{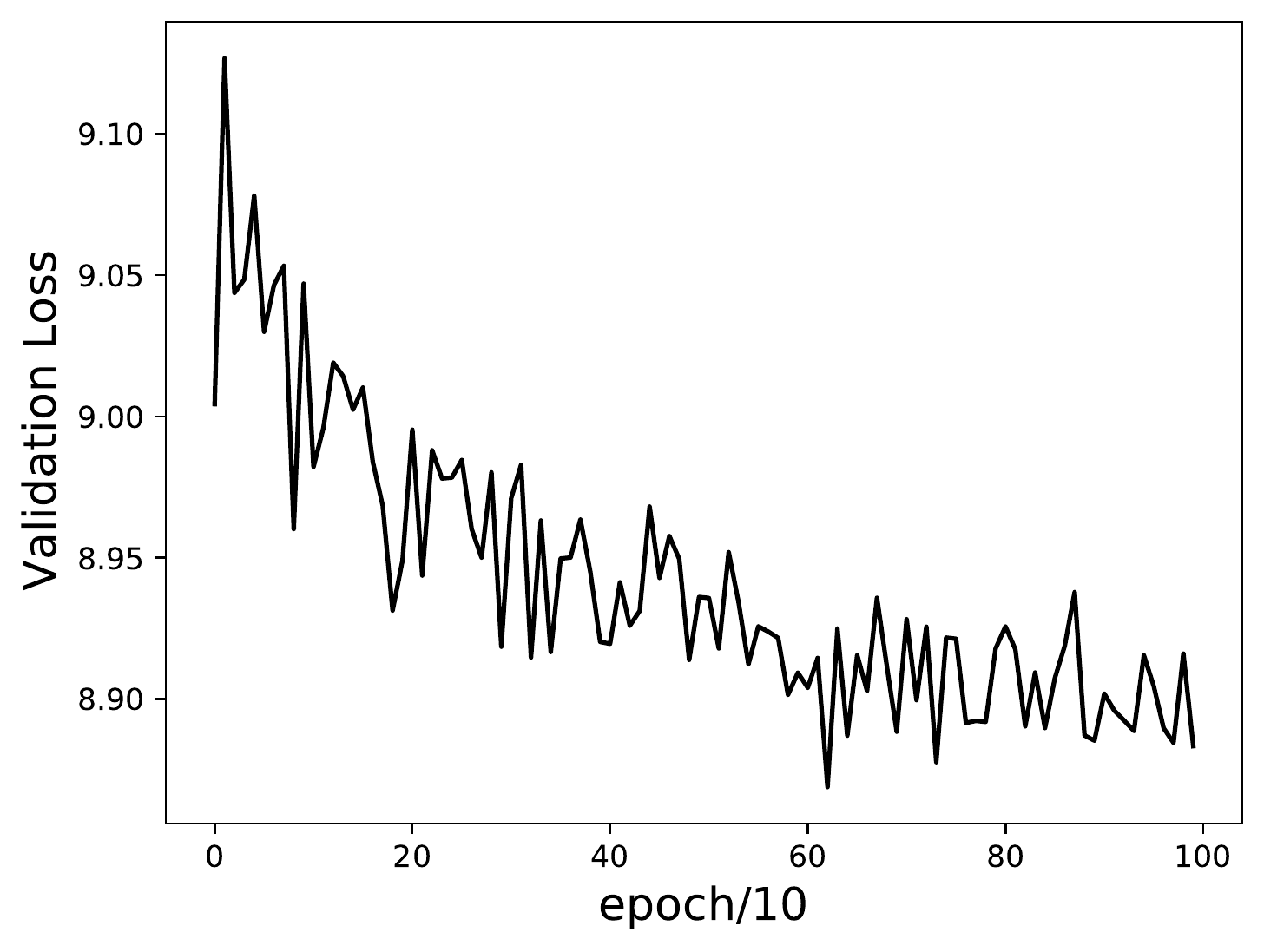}  
		\caption{}
		\label{fig:sub-first}
	\end{subfigure}
	\begin{subfigure}{.5\textwidth}
		\centering
		\includegraphics[width=5.4cm,height=4.1cm]{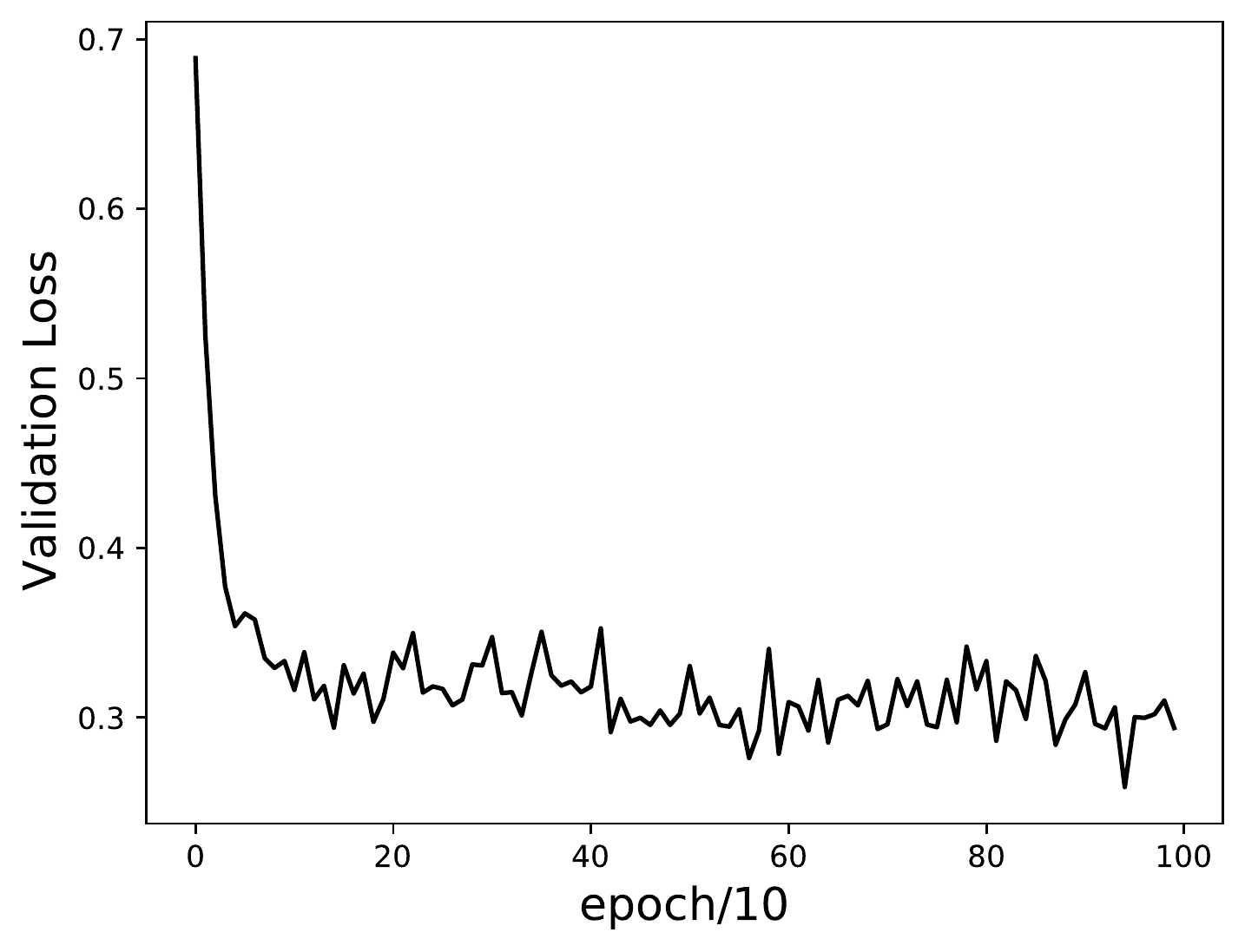}  
		\caption{}
		\label{fig:sub-second}
	\end{subfigure}	
	\newline	
	\begin{subfigure}{.5\textwidth}
		\centering
		\includegraphics[width=5.4cm,height=4.1cm]{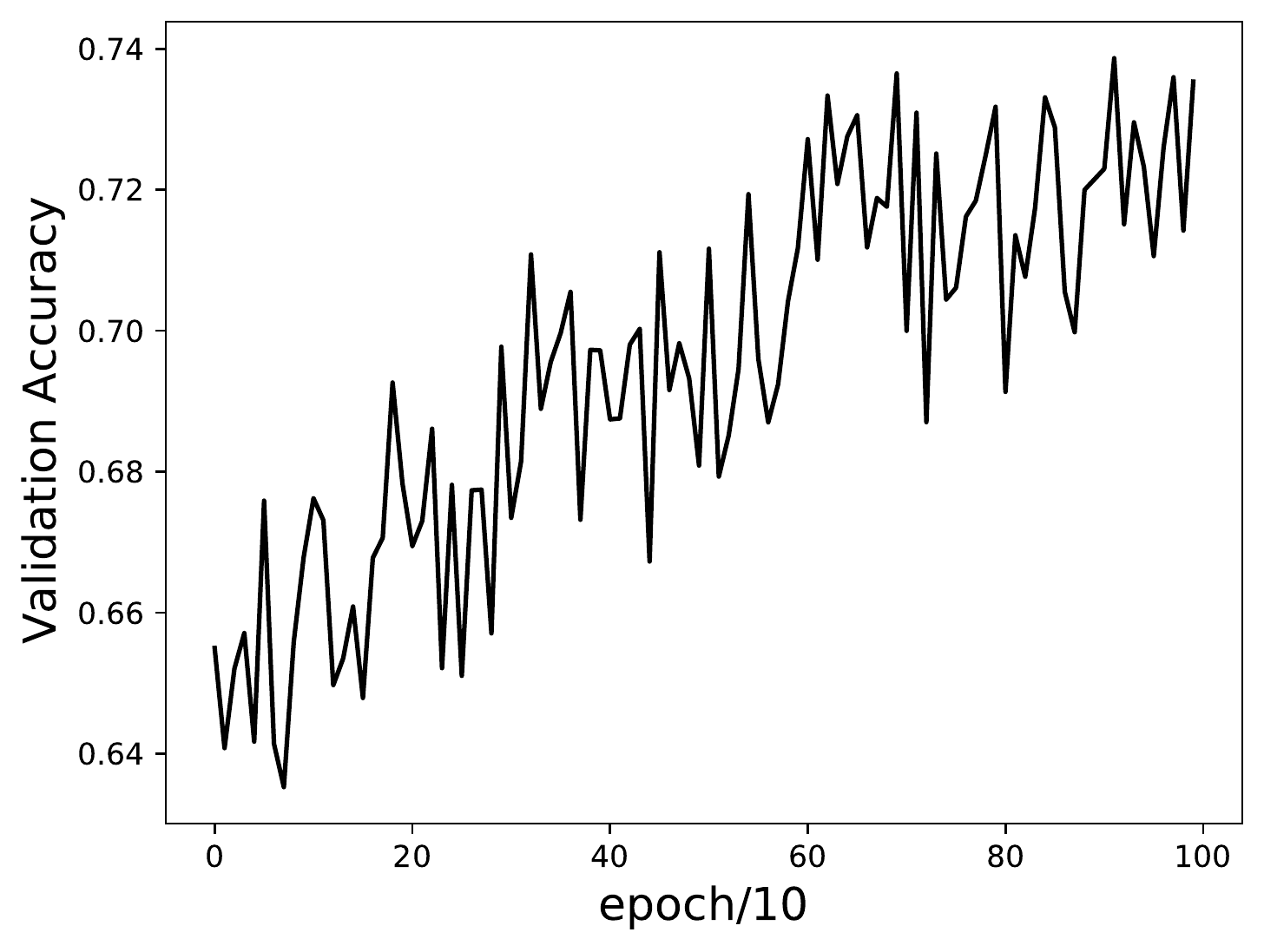}  
		\caption{}
		\label{fig:sub-third}
	\end{subfigure}
	\begin{subfigure}{.5\textwidth}
		\centering
		\includegraphics[width=5.4cm,height=4.1cm]{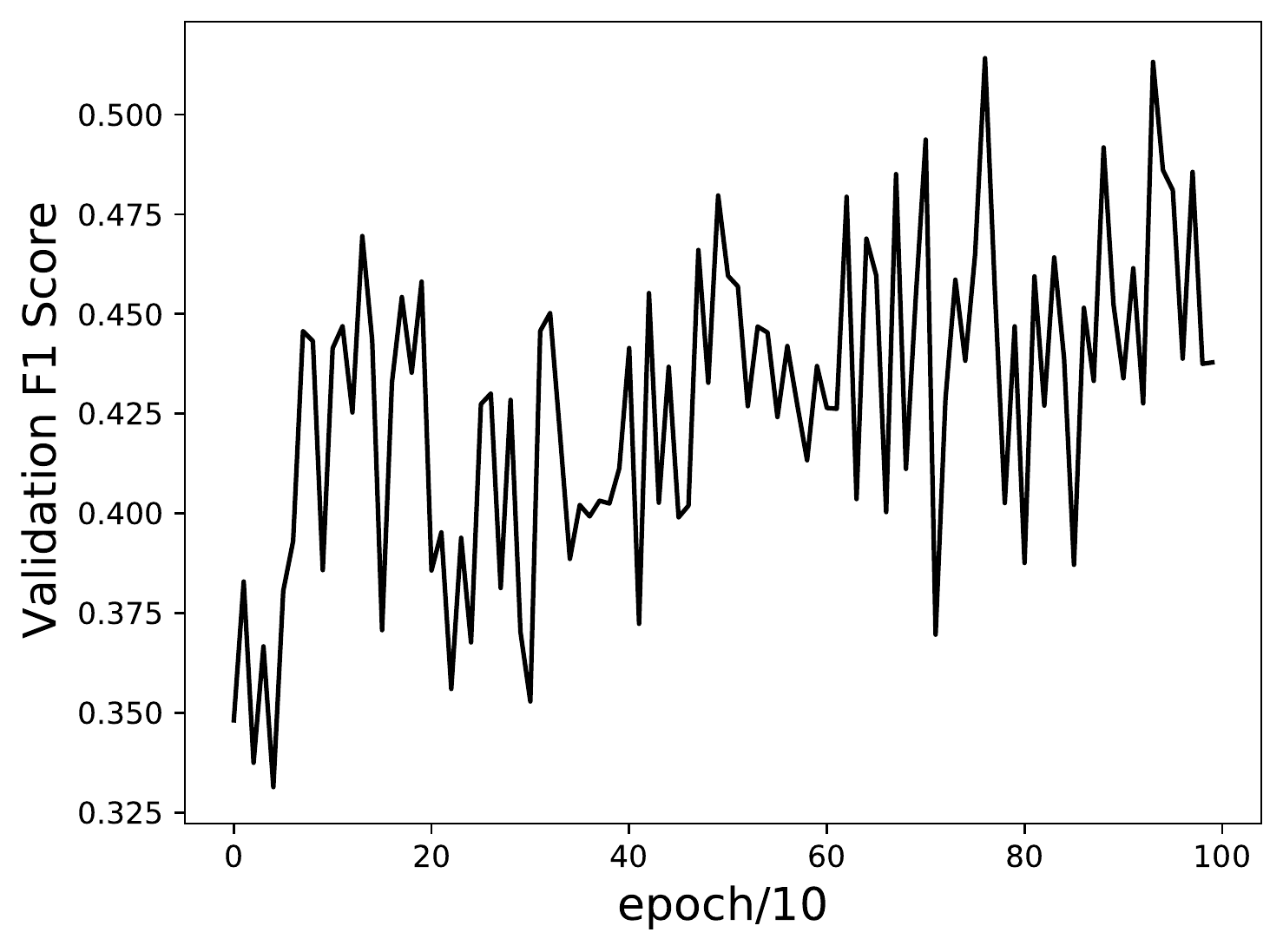}  
		\caption{}
		\label{fig:sub-fourth}
	\end{subfigure}
	\caption{Performance of GraphSAGE-MEAN: validation loss (a) and accuracy (c) of the unsupervised model and validation loss (b) and F1 score of the supervised model (d) calculated every 10 training epochs for a total number of 1000 epochs. Based on the performance development we set 1000 and 500 epochs to train unsupervised and supervised models, respectively.}
	\label{fig:fig3}
\end{figure}

For each unsupervised graph representation learning approach, the combination of learning rates and output dimensions result in 12 models where each model is trained for 1000 epochs (Figures \ref{fig:sub-first} and Figure \ref{fig:sub-third}) with a batch size of 1024 for the transductive and 2048 for the inductive tasks.

Mean values of micro-averaged F1-Scores for 1000 runs of the classifier on the representation vectors generated by each model are then calculated on the validation set and used to select our best performing model. Finally, micro-averaged F1-Score of 1000 runs of the classifier on the representation vectors generated by the best performing model using the test set is reported in table \ref{tab:tab2}.

The standard deviations of the F1-Scores tend to zero applying 1000 runs of the classifier on the representation vectors generated by each model.

\begin{table*}
	\begin{center}
		\begin{tabular}{|l| c| c| }
			\hline
			\multicolumn{3}{|c|}{\textbf{Parameter Settings}} \\
			\hline
			\textbf{Supervision}      & Unsupervised                             &  Supervised\\ 
			\textbf{Learning rate}    & $\{2e^{-8}, 2e^{-7}, 2e^{-6}, 2e^{-5}\}$ & $\{1e^{-4}, 1e^{-3}, 1e^{-2}\}$\\
			\textbf{Dimension}        & $\{64, 128, 256\}$                       & $\{64, 128, 256\}$\\
			\textbf{Epochs}           & 1000                                     & 500 \\
			\textbf{Sample Nodes}     & Layer1(9), Layer2(3)                     & Layer1(9), Layer2(3)  \\
			\textbf{Dropout}          & 0.1                                      & 0.1\\
			\textbf{Batch size}       & Task1(1024), Task2(2048)                 &   Task1(1024), Task2(2048)\\
			\hline
		\end{tabular}
		\caption {The table shows parameter settings for the experiments. \textit{Dimension} denotes the number of output dimensions of the representation vector at every depth $k$ of the recursion, which is set the same for both first and second layers. \textit{Task1} and \textit{Task2} represent transductive and inductive tasks, respectively.}
		\label{tab:tab3}
	\end{center}
\end{table*}

\paragraph{Supervised} 
In all supervised experiments, the exhaustive grid search is conducted over the learning rates of $\{1e^{-4}, 1e^{-3}, 1e^{-2}\}$ while the output dimensions of the representation vectors at every depth $k$ of the recursion are set to $\{64, 128, 256\}$. There are 9 and 3 neighbors sampled for aggregation in the first and second hop layers, respectively. There is a dropout rate of 0.1 used in this setting. Similar to our unsupervised settings, an exhaustive grid search is conducted to study the role of $\epsilon$ in GAIN (\ref{eq:6}).

For each supervised graph representation learning approach, all combinations of learning rates and output dimensions result in 9 models where each model is trained for 500 epochs (Figures \ref{fig:sub-second} and Figure \ref{fig:sub-fourth}) with a batch size of 1024 for the transductive and 2048 for the inductive tasks.

Mean values of micro-averaged F1-Scores of the class labels predicted by each model are then calculated for the validation set and used to select the best performing model when learning the task supervised. Finally, micro-averaged F1-Score of our best model on the test set is calculated and reported in table \ref{tab:tab2}.

\subsubsection{Comparison to RFN}\label{sec:rfn}
An RFN\cite{jepsen2020relational} consists of relational fusion layers where each layer is generated by a single layer perceptron and takes as input the node, edge, and between-edge representations from previous layer. 

To make a thorough comparison, the major differences of the method proposed in this study with RFN are mentioned in the following. In both cases road network graphs are extracted from OSMnx \cite{boeing2017osmnx} and used to generate input dataset, however RFN uses directed road network graphs of Danish cities while we make use of un-directed road network graphs of the Swedish cities.

RFN makes use of 3 raw node features represented by one-hot-encoded three dimensional vectors of city, rural, and summer cottage zones categories together with 10 raw edge features represented by one-hot encoded nine-dimensional vectors of road segment categories and one-dimensional length. Finally, they used 5 raw between-edge features represented by one-hot encoded four-dimensional vector of turn-direction together with one-dimensional turn angle. 

There are two sets of experiments conducted by RFN to address speed limit classification and speed limit estimation problems. The experiments are performed for the inductive task using a supervised binary classification on 4 versions of RFN based on the fusion and aggregation functions. We, on the other hand, address the road-type classification problem of multi-class input space using both supervised and unsupervised learning performed for the transductive as well as the inductive tasks.

To run our experiments with RFN, we utilized our graphs node, edge and between-edge features similar to other experiments performed in this study. The node features are represented by 2D node coordinates on the world map and the edge features are composed of 4 main components as described in section \ref{sec:feat}. Since between-edges represent common-nodes of the original graph, we, therefore, used the common-nodes attributes to generate between-edge features.

For comparison, we use the best performing model of \cite{jepsen2020relational}, RFN-A-I, which applies GAT \cite{velivckovic2017graph} aggregation and interactional fusion. Our supervised experiments with RFN-A-I are conducted using the soft-max cross-entropy loss function and ground-truth road type labels of multi-class input in our inductive task for 500 epochs. Similar to our other experiments, an exhaustive grid search is conducted over the learning rates of $\{1e^{-4}, 1e^{-3}, 1e^{-2}\}$ and the output dimensions of $\{64, 128, 256\}$ for the representation vectors at every depth $k$ of the recursion. The best performing model on the validation set is selected and tested on the test set. The results of 0.43\% accuracy for the transductive and 0.54\% accuracy for the inductive tasks are achieved, respectively.

\section{Discussion}

In this article, we propose a novel graph representation learning approach to address a road type classification problem. We also investigate state-of-the-art graph convolutional neural networks and their applications on road networks extracted from crowd-sourced open data. To this end, we designed 4 sets of experiments with a combination of transductive and inductive tasks when applying supervised and unsupervised learning.

To generate road segments representation vectors using GCN, GraphSAGE, GAT and GIN, we propose a transformation of the original graph into its line graph, where its nodes are composed of the original graph edges. Using this approach, we can perform downstream machine learning tasks on the road network as we demonstrate with the road type classification in our experiments.

To address the classification problem especially on large graphs, we proposed using a topological neighborhood composed of both local and global neighbors to train the graph-based loss function. This expands the node visibility of the graph structure and eventually improves performance. However, the extension of the topological neighborhood increases the amount of time and resources required to train the network. To address this limitation, we created the topological neighborhood of the node using one set of global neighbors only.

We developed a novel approach to aggregation, Graph Attention Isomorphism Network (GAIN), which based on our experiment results GAIN outperforms the state-of-the-art methods on unsupervised transductive task, and its performance is competitive in supervised transductive and inductive experiments.

For comparison, we conducted experiments with a number of state-of-the-art graph convolutional neural network methods proposed to aggregate the information of a node with its neighboring nodes using varying approaches. These are 7 state-of-the-art methods, including GCN \cite{kipf2016semi}, GraphSAGE (MEAN, MEANPOOL, MAXPOOL and LSTM) \cite{hamilton2017inductive}, GAT \cite{velivckovic2017graph} and GIN \cite{xu2018powerful}, trained using our road network data sets.

To make evaluations, we based our comparisons of the results with baseline performance achieved by random baseline and by applying only the raw features to the classifier. Baseline results are then compared with the performance of using representation vectors generated by different graph representation learning approaches shown in table \ref{tab:tab2}.

As shown in table \ref{tab:Transductive-Task}, all graph representation learning networks outperform the baseline on both supervised and unsupervised transductive task. However, GAIN outperforms the rest of networks in unsupervised experiments with 20\% improvement of performance compared to using raw features only. In supervised experiments, GAIN, GraphSAGE-LSTM and GraphSAGE-MEANPOOL perform the best with 37\% improvement over the classification results when using raw features only.

According to the results shown by table \ref{tab:Inductive-Task}, unsupervised inductive experiments are more successful than supervised ones. We hypothesize that using our proposed topological neighborhood composed of both local and global neighboring nodes, improves the classification of unsupervised experiments. However, the inductive experiment results shown in table \ref{tab:Inductive-Task}, also confirm that applying representation vectors trained by graph representation learning networks is superior to the baseline performance of road type classification problem.

In the unsupervised inductive task, GraphSAGE-MEANPOOL and GCN achieve the best results, which is 24\% above the baseline performance using raw features only. In supervised inductive experiments only GraphSAGE-MEAN and GAIN outperform baseline performance, with a maximum 22\% improvement to the baseline results when using raw features.

As presented by table \ref{tab:tab2}, road type classification task is quite challenging, which is mainly due to the input data characteristics such as imbalances of class distributions presented in Section \ref{sec:inputgt}, and figure \ref{fig:fig_}, upper image. Our motivation to merge some of the classes tries to address this problem, however, the problem still remains (see figure \ref{fig:fig_}, bottom image). Moreover, it generates identical roads with different features, which are labeled the same.

We also evaluate the performance of best RFN model, RFN-A-I \cite{jepsen2020relational} to address road type classification problem when learning supervised. Our results with RFN shows that it outperforms baseline performance on the inductive task but it fails on the transductive task. This could relate to the types of features used to describe nodes, edges and between-edges. Therefore, our results show that GAIN outperforms RFN \cite{jepsen2020relational} in supervised representation learning of road network graphs applying informative road segments attributes only.

\section{Conclusion}
In this paper, a novel approach to graph representation learning, GAIN, is proposed, which outperforms state-of-the-art methods shown in a wide set of experiments. GAIN is motivated through exploring the applications of different graph representation learning approaches in how to aggregate information available in the graph nodes vicinity using a designed topological neighborhood.

Highly representative attributes of the original graph edges, road segments, are applied to train graph vertices representation vectors using a line graph transformation. This addresses the problem of limited feature representation of original graph vertices (crossroads, junctions and intersections) since there are not sufficient features describing crossroads and intersections that are essential for road network representation.

To expand the graph nodes visual space, neighbor nodes are sampled from a topological neighborhood composed of both local and a set of associated global nodes through application of a random walk search mechanism proposed by algorithm \ref{alg:alg1}.

A main objective of this study is to present the application of our approach to unsupervised classification of real world road network graphs, since the complete annotation of the world is extremely expensive and it is also too cumbersome to re-adjust labeling of the roads, which are wrongly labeled. However, any classification task could be addressed using the same approach proposed in this paper.

Shown by our results the current work might be not ready for practical applications of road network graphs. It is still required to apply further tricks and regularization to improve its performance. As a future step, the challenges of road type classification could be addressed in a more realistic setting having more information of before and after segments as well as more informative road attributes.

\section{Acknowledgment}
This work was partially supported by the Wallenberg  AI,  Autonomous  Systems and Software Program (WASP) funded by the Knut and Alice Wallenberg Foundation and the Sweden’s innovation agency, Vinnova, through project iQDeep (number 2018-02700). The computations handling was enabled by resources provided by the Swedish National Infrastructure for Computing (SNIC), partially funded by the Swedish Research Council through grant agreement no. 2018-05973.

\newpage

\section*{References}

\bibliography{References}

\newpage

\begin{appendices}

\section{Appendix}	
Appendix with supplementary information regarding training and evaluation of the system. 

\subsection{Multi-head GAIN}
\label{sec:app1}

GAIN proposed by (\ref{eq:6}) can be easily extended using multiple heads for regularization, which might be interesting for other applications than the one presented in this article: 

\begin{equation}\label{eq:9}
\textup{{h}}_{v}^{k} = \textsc{{MLP}}^{k} ( \textup{{W}} \cdot ((1+\epsilon^k)\cdot \textup{{h}}_v^{k-1} + \sigma \sum_{m=1}^{M}\sum_{u \in N(v)} a_{m_{v, u}}^{k-1} \cdot \textup{{h}}_{u}^{\prime k-1})),
\end{equation}

where $\textup{{h}}_{u}^{\prime}=\textup{{W}}^{\prime}_m \cdot \textup{{h}}_{u}$ shows the linear transformation of node $u$ into a higher level feature space using weight matrix $\textup{{W}}^{\prime}_m$. The attention weight $a_{m_{v, u}}$ is given to the neighbor node $u\in N(v)$ where $m$ iterates over a number of attention heads used for regularization. $\textup{{W}}$ represents the shared weights matrices applied to optimize representation vectors of the sampled and the neighbor nodes. We also made experiments using (\ref{eq:9}) applying $M=1$ and we gained the results of unsupervised and supervised transductive task as $0.69$ and $0.80$, while for the unsupervised and supervised inductive task, we achieved the results as $0.60$ and $0.57$. This indicates that applying single head attention with different settings, (\ref{eq:6}) performs slightly better to when applying (\ref{eq:9}) in most cases of our experiments. Increasing the number of heads might improve the results for a more regularization, however it will be much more time consuming.

\end{appendices}

\end{document}